\documentclass[letterpaper, 10 pt, conference]{ieeeconf}
\IEEEoverridecommandlockouts                              

\usepackage{times}
\usepackage{multicol}
\usepackage[bookmarks=true]{hyperref}

\usepackage{algorithm}
\usepackage{algorithmicx}
\usepackage{algpseudocode}
\usepackage{graphicx}
\usepackage{amsfonts}
\usepackage{amsmath}
\usepackage{cleveref}
\usepackage{color}
\usepackage{multicol}
\usepackage{multirow}
\usepackage{subcaption}

\def\eg{\emph{e.g.}} 
\def\ie{\emph{i.e.}}

\newtheorem{assumption}{Assumption}

\begin{document}

\title{\LARGE \bf Towards Robust Deep Reinforcement Learning \\ against Environmental State Perturbation}

\author {
    Chenxu Wang and Huaping Liu\thanks{The authors are with the Department of Computer Science and Technology, Tsinghua University, Beijing, China. }
}

\maketitle

\begin{abstract}
Adversarial attacks and robustness in Deep Reinforcement Learning (DRL) have been widely studied in various threat models; however, few consider environmental state perturbations, which are natural in embodied scenarios.
To improve the robustness of DRL agents, we formulate the problem of environmental state perturbation, introducing a preliminary non-targeted attack method as a calibration adversary, and then propose a defense framework, named \textbf{B}oosted \textbf{A}dversarial \textbf{T}raining (BAT), which first tunes the agents via supervised learning to avoid catastrophic failure and subsequently adversarially trains the agent with reinforcement learning.
Extensive experimental results substantiate the vulnerability of mainstream agents under environmental state perturbations and the effectiveness of our proposed attack. The defense results demonstrate that while existing robust reinforcement learning algorithms may not be suitable, our BAT framework can significantly enhance the robustness of agents against environmental state perturbations across various situations.

\end{abstract}


\section{Introduction}
The safety and robustness of Deep Reinforcement Learning (DRL) have been receiving increasing attention and have been studied in various domains, such as perturbation on the observation \cite{sa-mdp, radial, zhang2021robust, liang2022efficient} or the action \cite{lee2020spatiotemporally}, data poisoning \cite{gunn2022adversarial, panagiota2020trojdrl}, adversarial policies \cite{gleave2019adversarial}, and multi-agent reinforcement learning \cite{lin2020robustness, guo2022towards}. 

\begin{figure}[t]
\centering
\includegraphics[width=0.5\textwidth]{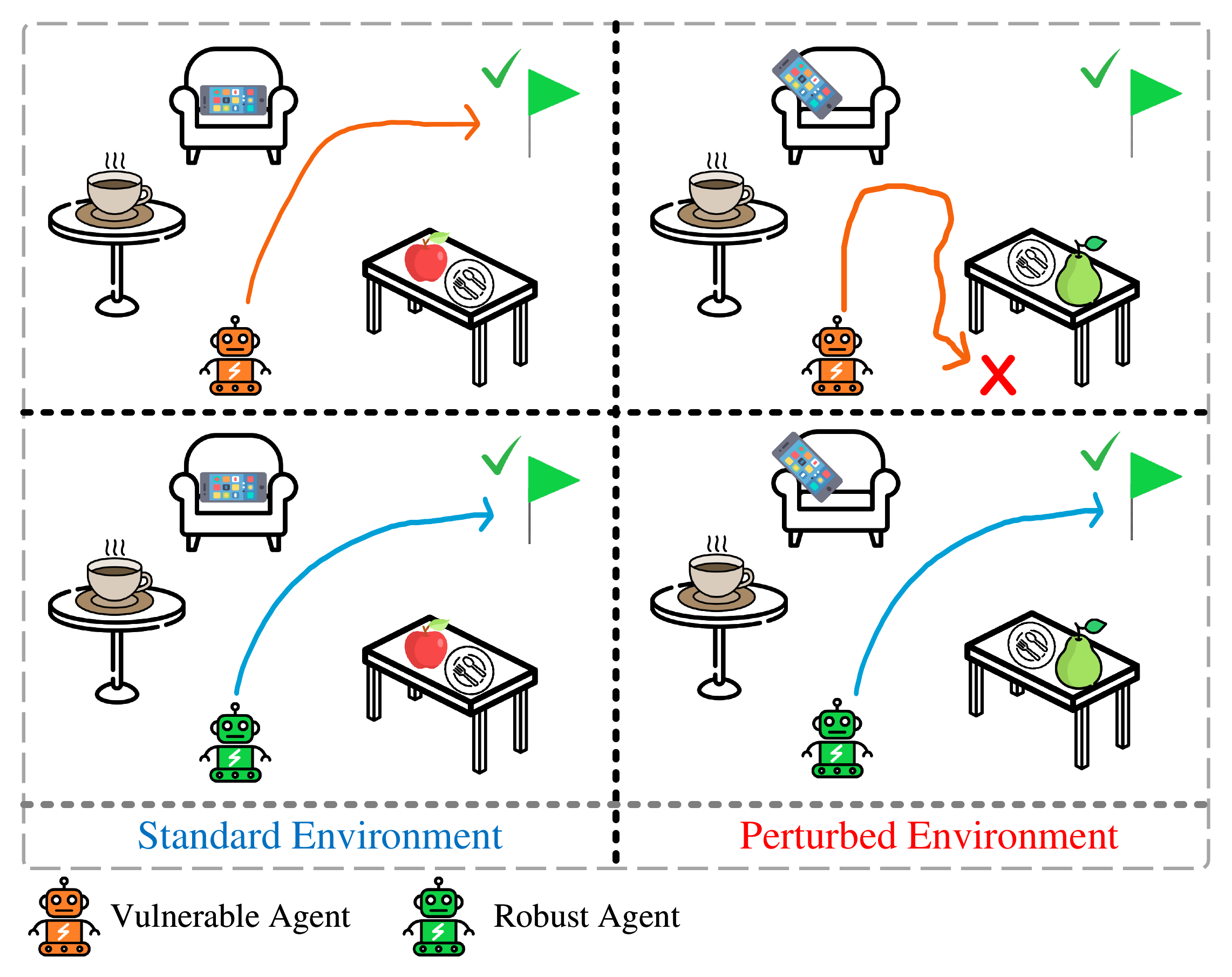}
\caption{An illustration of environmental state perturbation, where the arrangement of objects is changed. The perturbations, such as object movements, are assumed to be feasible, and even natural in the environment. While deliberately designed perturbations may incapacitate vulnerable DRL agents, robust agents are expected to defend against those perturbations. }
\label{fig: introduction}
\end{figure}

However, few works focus on the environmental robustness of the agent, which may be crucial for further applying DRL to robotic applications. In application, robots may be deployed in various environments that are different from the training one, with either unconscious or even malicious perturbations, such as placing or moving task-irrelevant objects. 
As exemplified in \Cref{fig: introduction}, we recognize the environmental state perturbations different from the traditional state perturbations in three characteristics: (1) The perturbations are only exerted onto the \textit{environmental} state with semantic restriction, instead of the whole state space under $L_p$-norm restriction. (2) The perturbations are static and only exerted on the initial state. (3) In consideration of real-world applications, the perturbed state should also be reachable from the standard initial state. 
The limitations ensure the alignment between the problem we studied and robotic application scenarios in reality, where the perturbation is practical and corresponding attacks are realizable.

To comprehensively study the environmental state perturbation, we first design a non-targeted white-box attack algorithm for both testing the existence of the vulnerability and serving as an adversary to examine the robustness of DRL agents. Then we propose a two-phase defense framework, which first kick-starts the agent via elaborated supervised learning, and subsequently fine-tunes the agent in the environment with generated adversarial initial states, namely \textbf{B}oosted \textbf{A}dversarial \textbf{T}raining (BAT).

We preliminarily inspect the environmental state perturbations and conduct both the attack and the defense experiments in a cooking game, Overcooked~\cite{bcp}, which has been found as a testbed for human-robot collaboration and collaborative DRL agents \cite{fcp, zhao2022coordination, yu2023learning, yan2023efficient}. Since the game simulates the interactions between agents and the environment, it is also applicable for the study of environmental perturbation, where the characters can be controlled by the same agent to meet the general DRL setting.

Experimental results show that our attack can significantly reduce the rewards of mainstream DRL agents, which substantiates the existence of environmental attacks and reveals the vulnerability of mainstream DRL agents under environmental state perturbations. On the other hand, our BAT
can greatly improve the robustness of DRL agents against such perturbations and even boost their performance in the pristine environment, clearly outperforming existing methods and the baselines.

The main contributions of this paper are:
\begin{itemize}
    \item We introduce the environmental state perturbation, a novel threat model where the saboteur perturbs the initial environmental state within the reachable set regarding the standard environment. 
    \item We formulate the attack problem and design an effective attack method that performs non-targeted attacks on policy-based agents. 
    \item We design the BAT framework, which includes a supervised kick-starting stage and a fine-tuning stage to enhance the robustness of agents to environmental perturbations. 
    \item We conduct comprehensive experiments for both attack and defense. The attack results show that the current DRL agents are vulnerable to environmental perturbations and their capabilities can be significantly decreased by our attack. The defense results show that our defense framework can effectively improve the resistance of the agents and outperform the baselines by a large margin. 
\end{itemize}
In summary, our work establishes a foundation to study environmental perturbations and may enlighten future DRL research.

\section{Related Work}
\label{sec: related_work}
\subsection{Adversarial attacks in DRL}
As a hot research topic, the adversarial vulnerability of DRL has been widely discussed. Inspired by the success of adversarial examples in Deep Neural Networks \cite{szegedy2013intriguing, su2019one}, researchers find that DRL is also vulnerable to small perturbations on inputs \cite{huang2017adversarial}. \cite{behzadan2017vulnerability} proposed a black box attack on Deep Q-networks (DQN) and validated the transferability of such an attack. By exploiting the characteristic of DRL, \cite{lin2017tactics} further proposed a strategically-timed attack to attack the agent at critical moments and an enchanting attack to induce the agent to a certain state. \cite{weng2020toward} proposed a model-based attack and extended the attacks into continuous domains. Universal adversarial perturbations that are static and agnostic to states are also studied to be powerful and efficient \cite{tekgul2022real}. \cite{buddareddygari2022targeted} noticed the importance of the physical availability of the attack and presented an algorithm to generate a static perturbation that can perform targeted attacks. Besides attacking the observations, \cite{schott2022improving} successfully attacked the agents by disturbing the dynamics of the environment. \cite{pan2022characterizing} studied various methods to improve the feasibility of attacks for application in reality and applied adversarial attacks to physical robots. \cite{lin2020robustness} performed adversarial attacks in cooperative multi-agent reinforcement learning (c-MARL) with a two-step attack that carefully deals with the characteristics of c-MARL. Similarly, \cite{guo2022towards} comprehensively tested the robustness of c-MARL agents by attacking the state, action, and reward. Beyond defeating the agent, adversarial methods can also be used for evaluating the agents as a way to find the worst case \cite{uesato2019rigorous}. 

However, most existing attacks use the threat model that perturbs the observation of the victim agent and constrains the perturbation by the $L_p$-norm, which requires real-time access to the agent. 
Although previous works have shown the effectiveness of perturbing the environment \cite{schott2022improving} and the initial state \cite{panda2018scalable}, they do not aim to feasible environmental attacks.
Starting from feasibility, our threat model restricts the environmental perturbations to be static and reachable in the original environment, which also makes the attack perceptually covert. Besides, the perturbations may also appear unintentionally due to the gap between the training environment and the production environment; thus, the attacks can also be regarded as an evaluation of the robustness of agents.

\subsection{Robust DRL}
Since the vulnerability of DRL has become a wide concern, studying and improving the robustness of DRL has become another spotlight topic. \cite{pinto2017robust} proposed RARL, which redefines the task as a zero-sum minimax problem and jointly trains a protagonist and an adversary. Similarly, training with an adversary embedded in environments has also been proven useful \cite{pattanaik2018robust}. Methods that optimize the agent against adversarial perturbations at the training stage have also been shown to be effective, such as SA-MDP \cite{sa-mdp}, RADIAL \cite{radial}, and WocaR \cite{liang2022efficient}. Another perspective is detecting attacks and defending actively \cite{tekgul2022real}. Besides, \cite{wu2022robust} introduced an adversarial curriculum learning framework to boost the robustness of agents. 
Other than perturbations on observations, \cite{schott2022improving} shows that training in environments with adversarial dynamics can also benefit the agent. RS-DQN \cite{fischer2019online} utilizes distillation to train a student DQN along with a standard DQN, thereby improving the robustness of the DQN. 
There are also works aiming to provide a guarantee or certification that the policy would not fail catastrophically when attacked by little perturbations \cite{lutjens2020certified, everett2021certifiable}. However, all the mentioned algorithms are designed under different threat models from ours, thus making it hard to resist our attack. To the best of our knowledge, no existing robust reinforcement learning algorithm is targeted at defending the environmental perturbations on the initial states.

\section{Preliminaries}

We concentrate on the embodied domain, where there exists an avatar of the agent in the environment, naturally splitting the state space into the agent state and the environmental state. We define such an embodied MDP as a tuple $<\mathcal{S}^{a}, \mathcal{S}^{e}, \mathcal{A}, P, R>$, where $\mathcal{S}^{a}$ is the space of agent states, $\mathcal{S}^{e}$ is the space of environmental states, $\mathcal{A}$ is the action space, $P: \mathcal{S}^{a} \times \mathcal{S}^{e} \times \mathcal{A} \times \mathcal{S}^{a} \times \mathcal{S}^{e} \mapsto [0, 1] $ is the transition dynamic, $R: \mathcal{S}^{a} \times \mathcal{S}^{e} \times \mathcal{A} \times \mathcal{S}^{a} \times \mathcal{S}^{e} \mapsto \mathbb{R} $ is a real-valued reward function.
For convenience, the joint state space is denoted as $\mathcal{S} = \mathcal{S}^{e} \times \mathcal{S}^{A}$.

We aim to improve DRL-based agents, where each agent is assumed to associate with a policy $\pi$.
For ease of expression, we introduce the notion $\pi_{opt}$ as a deterministic version of $\pi$, and denote $a^* = \pi_{opt}(s^a, s^e) = \max_{a \in \mathcal{A}} \pi(a | s^a, s^e)$ as the deterministic action output, where $s^a \in \mathcal{S}^{a}$ is the agent state, and $s^e \in \mathcal{S}^{e}$ is the environmental state.
During the reinforcement learning process, a policy $\pi$ is trained to maximize the expected reward $\rho(\pi, s^{a}_0, s^{e}_0) = \mathbb{E}_{\tau} \sum_{t} R(s^{e}_t, s^{e}_t, a_t)$, where $\tau$ is the trajectory sampled following agent policy $\pi$ under the initial agent state $s^{e}_t$ and the initial environmental state $s^{e}_0$, and $t$ stands for the time step.

The attack problem aims to find out an initial environmental state $\widehat{s^{e}_0} $ that minimizes the expected reward, formulated as:
\begin{gather}
    \min_{\widehat{s^{e}_0} \in \widehat{\mathcal{S}^{e}} } \rho(\pi, \widehat{s^{e}_0}, s^{a}_0), \\
    {\rm{s.t.~~~~}} \mathbb{D}(s^{e}_0, \widehat{s^{e}_0}) \leq \epsilon. \notag
\end{gather}
Where $\mathbb{D}$ is a function that measures the semantic distance between states and $\epsilon$ is the threshold. It is noteworthy that $\mathbb{D}$ may not be a conventional distance metric on the observation space such as $L_p$-norm, but a semantic metric that is specified by the experiment domain instead. 
Similarly, the defense problem is defined as training a robust agent against the environmental state perturbations, which is a corresponding max-min problem, formulated as:
\begin{gather}
    \max_{\pi} \min_{\widehat{s^{e}_0} \in \widehat{\mathcal{S}^{e}} } \rho(\pi, \widehat{s^{e}_0}, s^{a}_0), \\
    {\rm{s.t.~~~~}} \mathbb{D}(s^{e}_0, \widehat{s^{e}_0}) \leq \epsilon. \notag
\end{gather}

However, simulating in the environment to get the reward $\rho$ is time-consuming, and thus exhaustive search is not applicable. Therefore, we design an attack algorithm with an alternative objective function as an approximate solution and then develop a defense algorithm to train robust agents, both will be introduced in \Cref{section: Methodology}. 

Before diving into the algorithms, we make a perturbation invariance assumption as follows:
\begin{assumption}
    Let $s^{e}_t$ denote the states in trajectories collected following policy $pi$ under the standard setting and $\widehat{s^{e}_t}$ denotes those under the perturbed initial environmental state $\widehat{s^{e}_0}$, then for successful attacks where $\rho(\pi, s^{a}_0, \widehat{s^{e}_0}) << \rho(\pi, s^{a}_0, s^{e}_0)$, we assume $|\widehat{s^{e}_t} - s^{e}_t| \geq |\widehat{s^{e}_0} - s^{e}_0|$.
\end{assumption}
This assumption supposes that perturbations on the initial state will not be eliminated if the attack succeeds, and implies the perturbations do not directly hinder the agents. Such an intuition is utilized in both our attack and the defense. Besides, our defense algorithm assumes the policy using the Softmax function to map the outputs into action probabilities, nevertheless, its framework can be generalized to other policy architectures.


\section{Methodology}
\label{section: Methodology}
\begin{figure*}[t]
\centering
\includegraphics[width=0.9\textwidth]{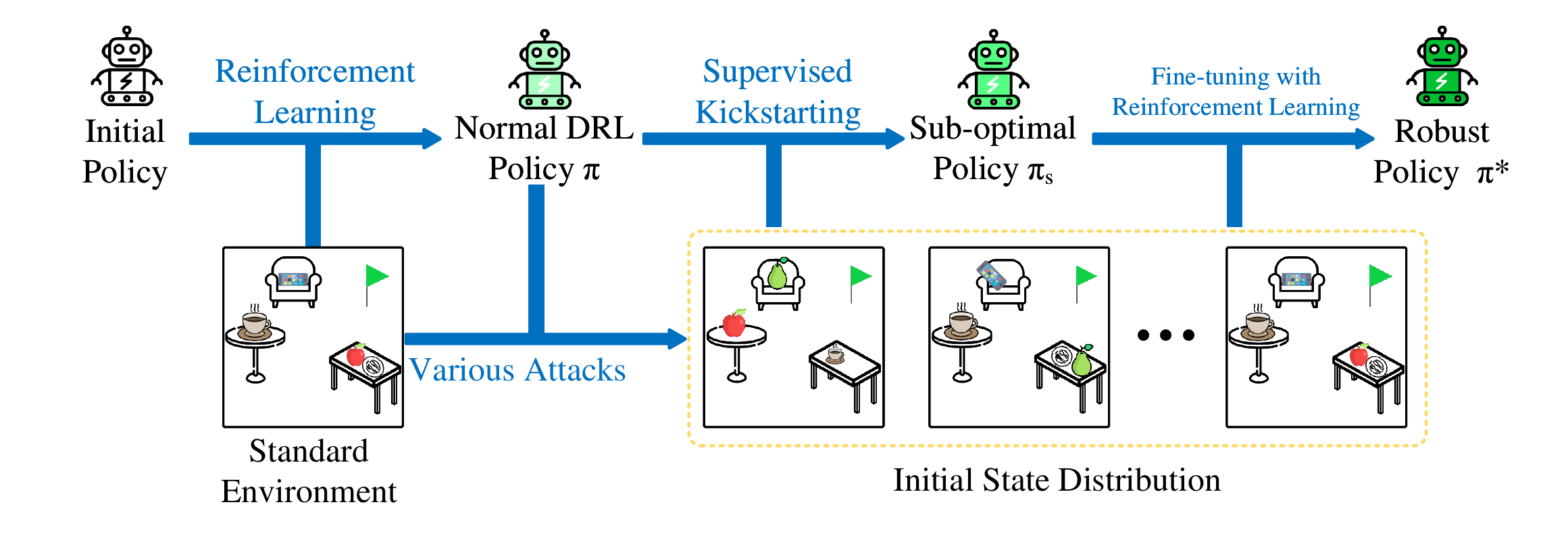}
\caption{An illustration of the entire procedure of our BAT framework, which utilizes adversarial perturbations to adaptive defense.}
\label{fig: defense_architecture}
\end{figure*}

This section introduces our approach for training robust DRL agents against environmental state perturbation as illustrated in \Cref{fig: defense_architecture}, including a preliminary attack algorithm for both calibrating the robustness of DRL agent and providing adversarial examples, following by our BAT framework for improving the robustness of DRL agents utilizing the generated adversarial environmental states.

\subsection{Attacks}
Similar to previous works, we generate adversarial examples against trajectories collected from the original environment, keeping the computational efficiency while acquiescing to ignore the possible chain reaction caused by the perturbations. 
We observe that environmental perturbations would cause more capability generalization failure rather than goal misgeneralization introduced by \cite{di2022goal} and seldom interfere with the original trajectories of agents.
Therefore, we aim to perform a non-targeted attack that induces the agent to deviate from its original actions and define the objective of our attack algorithm as:

\begin{align}
    & \max_{\widehat{s^{e}_0} \in \widehat{\mathcal{S}^e} } ~~  - \mathbb{E}_{\tau} \left[\pi(a_t^* |s^a_t, \widehat {s^e_t})\right], \\
    & {\rm{s.t.~~~~}} \mathbb{D}(s^{e}_0, \widehat{s^{e}_0}) \leq \epsilon. \notag
\end{align}

where $\tau$ denotes the trajectory, $a^*_t = \pi_{opt}(s^{a}_t, s^{e}_t)$ denotes the optimal action suggested by $\pi$, $\widehat {s^e_t}$ is the estimated environmental state after perturbations, and $\epsilon$ is the limitation of the perturbation distance.

To further reduce the computational cost and make the algorithm scalable, we approximately compute the significance of perturbation with the first-order derivation to avoid tremendous calls of the policy network.
Then, with collected trajectories $\tau$, we have a computationally acceptable function to estimate the attack effectiveness given an adversarial state $\widehat{s^{e}_0}$:
\begin{align}
\label{form: Objective}
J(\hat{s}^{e}_0) = \sum_{(s^{a}_t, s^{e}_t, a^*_t) \in \tau} \left[\frac{\partial \pi(a^*_t | s^{a}_t, s^{e}_t)}{\partial s^{e}_t}(s^e_0 - \hat{s}^{e}_0) \right].
\end{align}
Where $\widehat{s^{e}_0} - s^{e}_0$ is the state difference introduced by the environmental state perturbation on the initial state, being assumed time-invariant.

The optimization of \Cref{form: Objective} highly depends on the environment. For environments with a discrete state space, gradient-based optimization may not be applicable. Generally, the problem can be solved via generic algorithms by taking the $J(\hat{s}^{e}_0)$ in \Cref{form: Objective} as the fitness function and let $\mathcal{S}^{e}$ be the search space, while the solution may be further simplified regarding the characteristics of the environment as specified in \Cref{subsec: Environment}.
From the insight that out-of-distribution perturbations incapacitate agents, we filter out perturbations that have been observed in the collected trajectories over a frequency threshold $p_{freq}$.

\subsection{Defenses}
As the results will be discussed later in \Cref{sec: Experiments}, the existing DRL agents expose their vulnerability to environmental attacks, which necessitates the study of robust training algorithms. A straightforward idea is adversarial training, \ie, to fine-tune the DRL agents on the adversarial initial states. However, this may not practically work when the agents are severely disabled by the attacks, since the reinforcement learning process requires learning from trajectories with rewards, while the agent may fail to get any rewards in such cases and consequently fail the adversarial training.
To address such a problem, we introduce our BAT framework as shown in \Cref{fig: defense_architecture}, which first utilizes supervised learning in a teacher-forcing way as a kick-starting and then fine-tuning the agent with reinforcement learning under a mixed distribution of initial states.

The kick-starting stage aims to avoid the catastrophic failure caused by the environmental attack, where we train a start policy $\pi_{s}$ that maintains a fundamental capability in the perturbed environments. Carrying on the insight that environmental perturbations have limited direct influence on the agent, we force the $\pi_{s}$ to act in the perturbed environments similarly to the pristine one. 

In this paper, we conduct a case study on the actor-critic policy, which is a relatively complicated case, and the method can be generalized to value-based or policy-based algorithms. Typically, in the case of actor-critic, the policy $\pi$ that computes the actions, and a critic
$V: \mathcal{S}^{a} \times \mathcal{S}^{e} \mapsto \mathbb{R}$ estimates the value of states, while they may share some parameters. 

We hope the new policy $\pi_s$ can keep the learned skills in the pristine environment; therefore, we constrain the agent close to its original version by using the following loss functions:

\begin{align}
    \mathcal{L}_{o} = & \mathbb{E}_{\tau} \left[ KL(\pi(\cdot | s^{a}, s^{e}) || \pi_s(\cdot | s^{a}, s^{e}) \right] \notag \\
    & + |V(s^{a}, s^{e}) - V_s(s^{a}, s^{e})|,
\end{align}

where $KL$ stands for the Kullback–Leibler divergence, $V_s$ is the new value function.

We also expect the agent's behaviors in the perturbed environments to be similar to those in the original environment, as a feasible start for fine-tuning. However, the outputs are not necessarily the same, which may make the agent simply ignore the environmental changes.
Inspired by the Knowledge Distillation \cite{hinton2015distilling, papernot2016distillation}, we set a higher temperature $T$ for the Softmax function of the policy to acquire such labels for supervised learning. The policy with temperature $T$ is defined as:
\begin{gather}
    \pi^{T}(a | \cdot ) = \left[ \frac{e^{z_a / T}}{\sum_{a \in \mathcal{A}} e^{z_{a} / T}} \right] ,
\end{gather}

where $z \in \mathbb{R}^{|\mathcal{A}|}$ is the output of the last layer. The original policy $\pi$ is a special case where $T=1$. Similarly, the values of disturbed states are expected to be close to the original ones, while not forced to be the same. Therefore, we loosen the L1 loss of value function to achieve this purpose. The loss function on the disturbed data can be written as:
\begin{align}
    \mathcal{L}_{p} = & \mathbb{E}_{s \sim \tau} [ KL(\pi^{T}( \cdot | s^{a}, s^{e}) || \pi_s( \cdot | \hat{s^{a}, s^{e}}))  \notag \\ 
    & + max(|V(s^{a}, s^{e}) - V_s(\hat{s^{a}, s^{e}})| - \alpha|V(s^{a}, s^{e})|, 0) ],
\end{align}

where $\hat{s}$ denotes the estimated perturbed state, $\alpha$ is a hyperparameter that controls the proximity threshold between the original values and disturbed ones. 
The loss terms above are then combined as $\mathcal{L} = \mathcal{L}_{o} + \beta \mathcal{L}_{p}$, where $\beta$ is a hyperparameter that controls the weights of each part. We then minimize loss $\mathcal{L}$ via supervised learning to get the $\pi_s$. In our practice, the perturbed initial states include adversarial initial states generated by the attack algorithm and random states in the feasible set. Unlike the popular practice in knowledge distillation and student networks, our fine-tuning starts from the trained policy. It is worth emphasizing that the supervised learning phase does not immediately lead to a robust policy since it is designed to induce the policy to a good start of consequent fine-tuning.

We then fine-tune the policy $\pi_s$ in the environment with a distribution of initial environmental states $S^e_0$ via reinforcement learning algorithms, where the objective can be described as:
\begin{gather}
    \max_{\pi} \mathbb{E}_{s^e_0 \sim S^e_0} \left[ \rho(\pi, s^a_0, s^e_0) \right] .
\end{gather}

Typically, the initial state distribution $S^e_0$ consists of the original initial state and the perturbed initial states used in kick-starting. It is noteworthy that although our defense method requires extra training, the fine-tuning process usually takes less time than training a standard DRL agent; thus, our method does not bring an excess computational burden.

\begin{figure*}[t]
\centering

\includegraphics[width=0.95\textwidth]{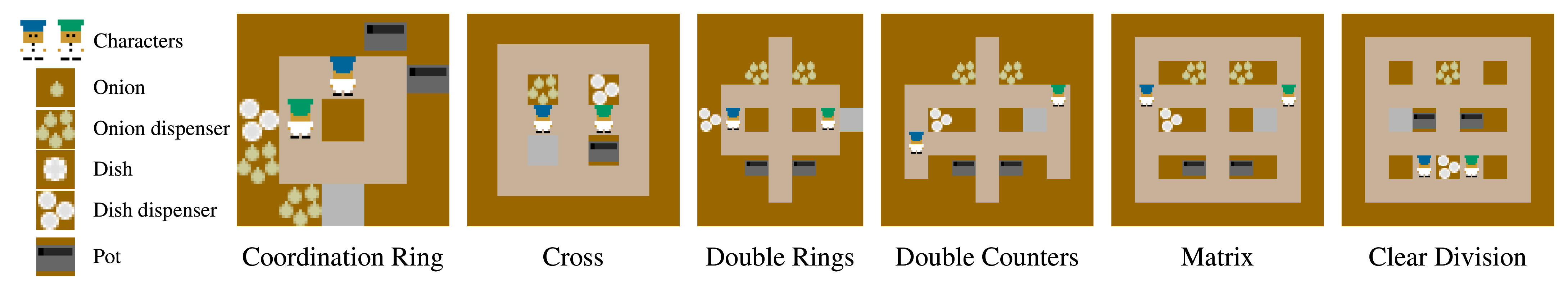}
\caption{Illustration of the icons and layouts in the Overcooked environment.}
\label{fig: adversarial examples}
\end{figure*}

\section{Experiments}
\label{sec: Experiments}

\subsection{Testing environment}
\label{subsec: Environment}
Though environmental perturbations may be ubiquitous in the real world, it is hard to model in simulations or games. The mainstream benchmarks for RL may not explicitly incorporate the interaction between agents and the environment, \eg, Mujoco \cite{todorov2012mujoco} and ProcGen \cite{cobbe2020leveraging}. Therefore, we test DRL algorithms and all our methods in the Overcooked environment \cite{bcp}, which has an interaction mechanism and was originally developed as a cooperative environment for studying multi-agent collaboration and human-AI collaboration. To fit the general single-agent DRL setting, the environment is used in self-play mode, \ie, a single agent controls the two characters. Agents are trained to cook soup and serve it at certain locations. Typically, the agent needs to: put onions into a pot to make soup; dish out the soup once it is ready; and deliver it to a serving location. The action space of each character includes 6 actions: $wait$, $move \{up, down, left, right\}$, and $interact$. The agent can interact with environments, such as putting or taking objects to or from counters.
Our experiments cover 6 layouts as shown in \Cref{fig: adversarial examples}, which are of various difficulties so that the effectiveness of algorithms can be comprehensively tested. 

Since all the counters are empty in the standard initial state, the removal or movement of an object is not applicable in our experiments, leaving the following possible categories of
unit perturbation: 
\begin{enumerate}
    \item Putting an onion on a reachable empty counter;
    \item Putting a dish on a reachable empty counter;
    \item Putting any onions in an empty pot.
\end{enumerate}
The perceptive distance between environmental states $\mathbb{D}(s^e, \widehat{s^e})$ is defined as the minimal number of unit perturbations to transform $s$ into $\hat{s}$. 

The Overcooked simulator provides a lossless input in the space $\mathbb{R}^{26*w*h}$ corresponding to the state, where $w$ and $h$ are the width and height of the layout. Fortunately, each unit perturbation listed above independently exerts influence on the observation. 
Therefore, instead of heuristic searching, we enumeratively calculate the attack objective for each unit perturbation and combine the ones with the highest estimated effect.

\begin{figure*}[t]
\centering
\includegraphics[width=\textwidth]{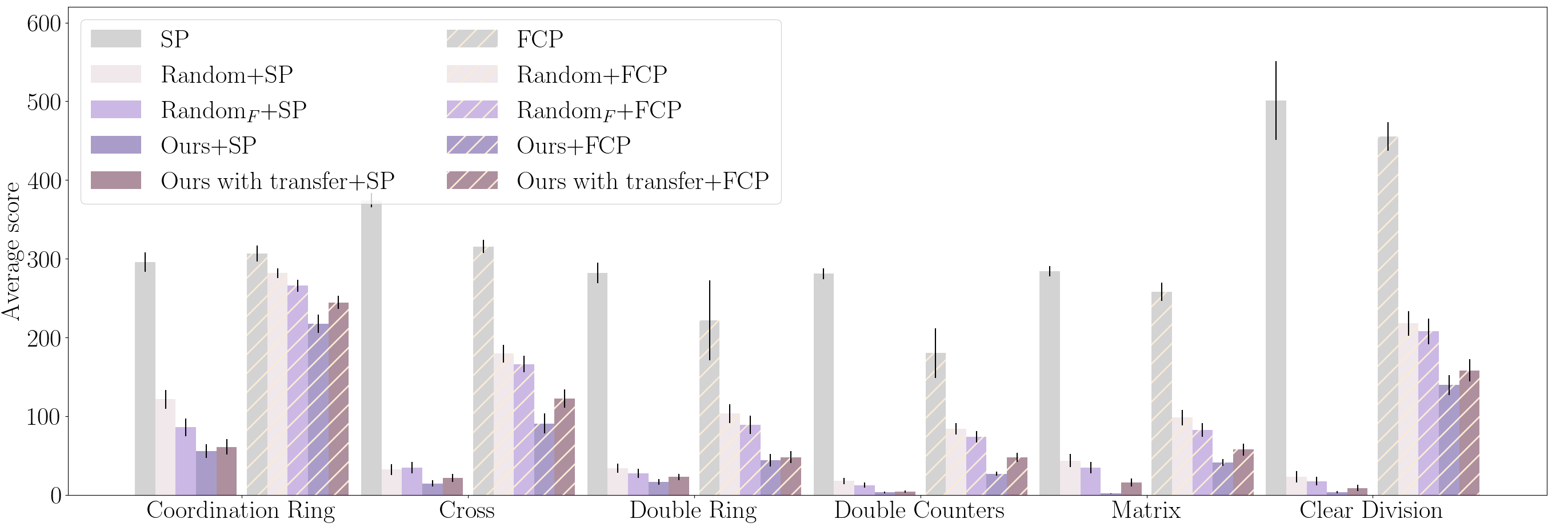}
\caption{The mean scores and standard error across agents for SP and FCP.}
\label{fig: results}
\end{figure*}

\begin{figure}[t]
\centering
\includegraphics[width=0.48\textwidth]{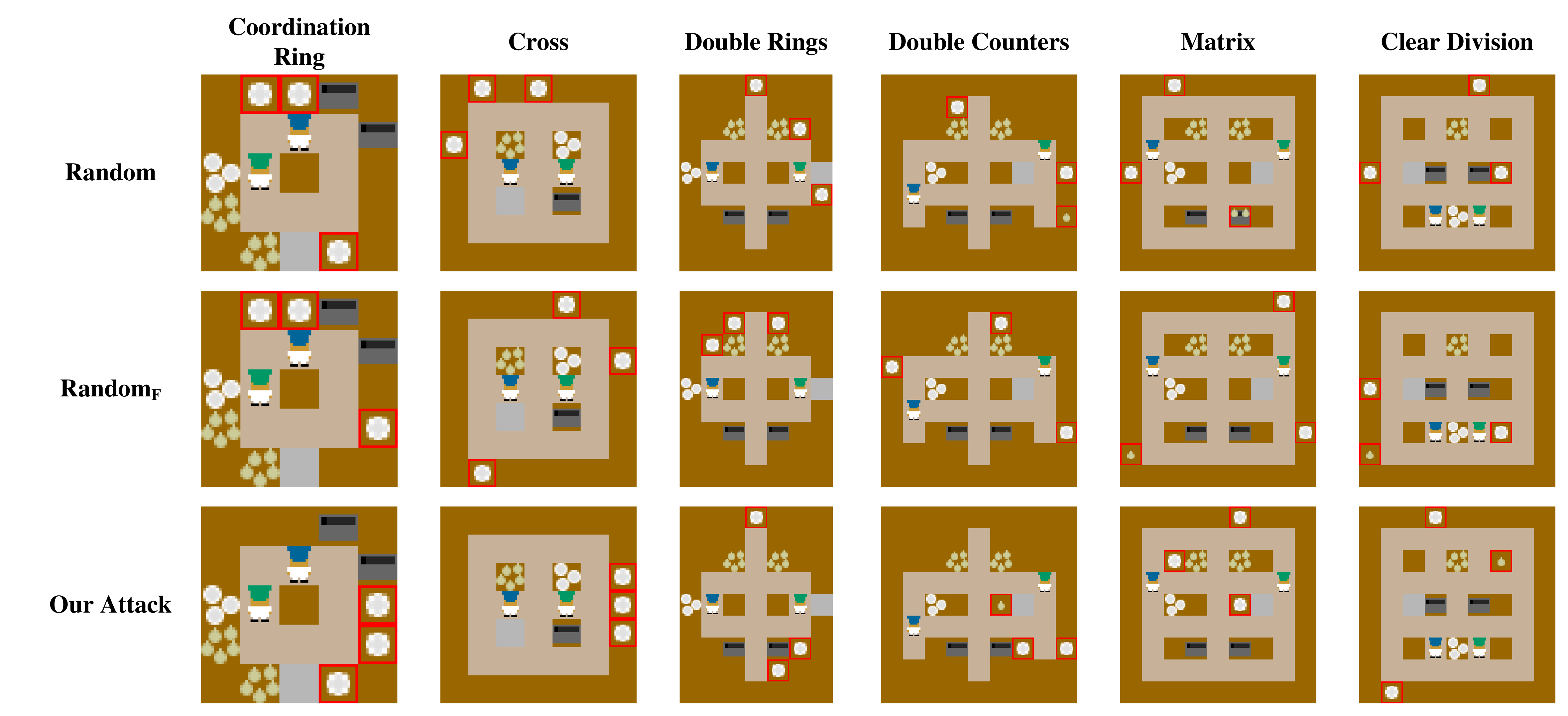}
\caption{Environmental state perturbations generated by all methods that reduce the reward of SP agents the most.}
\label{fig: examples}
\end{figure}

\subsection{Attack}
\subsubsection{Experimental setup}
\label{subsubsec: Attack Settings}
We conduct attack experiments on two popular DRL agents, Self-Play (SP) and Fictitious Co-play (FCP) \cite{fcp}. In SP, the policy controls the two agents at the same time and is optimized with the joint trajectories. FCP is a strong algorithm designed for zero-shot coordination and human-AI collaboration, which trains the agent with a diversified group of pre-trained partners. For FCP, we train 4 partners with different seeds to form the partner pool and select 3 checkpoints with different levels of abilities for each partner.
All agents are trained with the PPO algorithm \cite{schulman2017proximal} for $1e7$ environment steps and have the same network architecture as in \cite{bcp}. For both DRL algorithms, we train 5 independent agents. 

To comprehensively evaluate our attack algorithm, we take the following methods into comparison as baselines:

\begin{itemize}
    \item \textbf{Random}. The adversarial initial states are randomly selected from the feasible set. 
    \item \textbf{Random$_F$}. Randomly selected adversarial initial states after filtering out the perturbations that have appeared in the collected trajectories over a certain frequency. The setting of $p_{freq}$ is the same as our attack algorithm.
    \item \textbf{Attack with transfer}. To evaluate the transferability of our attack, we test the performance of agents with perturbations generated by attacking other agents. 
\end{itemize}

For all attacks, the perturbation limitation $\epsilon$ is set to $3$. The number of output adversarial initial states $k$ of our attack algorithm is set to $10$. To reduce the randomness, the $k$ of the random baselines was set to $40$. For the evaluation of transferability, we use all the $40$ adversarial initial states generated by attacking the other 4 agents.
For each adversarial initial state, we run 100 games with 800 environment steps per game. We report the average score across all adversarial initial states and all 5 agents for each category. The evaluation settings remain the same throughout the experiment.

\subsubsection{Attack results}

The effectiveness of attacks on initial states can be revealed by the decrease in the average rewards. As the results are shown in \Cref{fig: results}, our attack significantly reduces the rewards of both agents in all layouts and clearly outperforms the random baselines. In some cases, the rewards even go down to near zero, such as SP in \textit{Double Counters} and \textit{Matrix}, indicating completely incapacitated agents. Such results demonstrate the effectiveness of our attack algorithm. Although the transferred attack shows less effectiveness than the white-box attack, it still consistently outperforms the random baselines, indicating the transferability of our attack. We also notice that the Random$_F$ is stronger than the Random, which testifies to our insight that out-of-distribution states may more effectively incur failures.
Furthermore, we find that even random perturbations can considerably decrease the rewards on all layouts, though they can be unintentional and are agnostic to the DRL agent, exposing that the mainstream DRL algorithms lack robustness to environmental state perturbations. Even without a malicious adversary, the agent may fail due to an unexpected misalignment between the training and testing environments, underlining the need to study environmental robustness for DRL agents.

Comparing the two DRL algorithms, FCP agents perform significantly better in resisting the attacks, although they may have lower scores in the vanilla environment. Recall that the FCP agents are trained with a set of diversified partners and have explored more states in their training data, this result fits the common perspective that the robustness of agents benefits from training with more diverse data. Moreover, comparing across layouts, we find the attack is less effective in simpler layouts such as \textit{Coordination Ring}, and is much more significant in complex layouts. Such a phenomenon suggests that the success of attacks comes from unknown and unreasonable behaviors when the agents encounter out-of-distribution observations, and naturally leads to an insight that agents may benefit from training in more diversified environments and adversarial training. 

Some strong adversarial examples of environmental state perturbations within an attack budget $\epsilon = 3$ are illustrated in \Cref{fig: examples}. Although the perturbations are reachable in the game, seemingly casually placing objects on some counters, they can significantly influence the agent. Though the power of perturbations is hardly visually distinguishable for humans, deliberately chosen perturbations can more effectively incapacitate the agent, further substantiating the effectiveness of our proposed attack algorithm.

\begin{table*}[htb]
    \centering
    \caption{Quantitative Defense results (average scores with standard errors)}\label{tab: dfd_result}
    \scalebox{1}{
    \begin{tabular}{cccccccc}
    \hline
    Method & Attack & Coord. Ring  & Cross & Doub. Rings & Doub. Coun. & Matrix & Clear Div.  \\

    \hline
    \multirow{3}{*}{Extra SP} & No attack & 341.8 $\pm$ 16.2 & 399.5 $\pm$ 11.6 & 325.2 $\pm$ 7.8 & 304.7 $\pm$ 7.3 & 320.6 $\pm$ 13.2 & 549.5 $\pm$ 41.1  \\

    & Random & 88.4 $\pm$ 12.8 & 36.1 $\pm$ 8.2 & 54.1 $\pm$ 8.4 & 29.8 $\pm$ 6.6 & 46.9 $\pm$ 8.2 & 23.5 $\pm$ 6.2  \\
    & Our attack &  36.8 $\pm$ 6.3 & 5.0 $\pm$ 1.2 & 39.4 $\pm$ 6.3 & 9.3 $\pm$ 2.6 & 10.8 $\pm$ 2.9 & 4.9 $\pm$ 1.0  \\

    \hline
    \multirow{3}{*}{Extra FCP} & No attack & 365.6 $\pm$ 4.2 & 399.2 $\pm$ 7.4 & 327.6 $\pm$ 14.6 & 324.4 $\pm$ 12.1 & 307.6 $\pm$ 8.0 & 631.0 $\pm$ 12.9  \\

    & Random &  234.7 $\pm$ 13.9 & 75.8 $\pm$ 11.0 & 44.5 $\pm$ 6.9 & 37.8 $\pm$ 6.6 & 16.8 $\pm$ 4.8 & 45.6 $\pm$ 12.7  \\

    & Our attack & 121.3 $\pm$ 12.9 & 19.4 $\pm$ 2.8 & 37.7 $\pm$ 9.6 & 6.6 $\pm$ 1.8 & 1.0 $\pm$ 0.3 & 5.1 $\pm$ 1.2  \\

    \hline
    \multirow{3}{*}{RADIAL} & No attack &  273.6 $\pm$ 9.6 & 287.6 $\pm$ 7.1 & 204.6 $\pm$ 6.7 & 218.4 $\pm$ 8.2 & 202.2 $\pm$ 8.7 & 303.5 $\pm$ 13.1  \\

    & Random &  138.8 $\pm$ 10.1 & 47.9 $\pm$ 10.4 & 43.9 $\pm$ 7.3 & 75.3 $\pm$ 9.3 & 40.4 $\pm$ 7.5 & 39.7 $\pm$ 8.7  \\

    & Our attack & 103.1 $\pm$ 9.1 & 51.1 $\pm$ 7.2 & 15.1 $\pm$ 2.8 & 53.3 $\pm$ 7.3 & 14.7 $\pm$ 3.7 & 25.3 $\pm$ 5.6  \\

    \hline
    \multirow{3}{*}{Div. Start} & No attack & 311.7 $\pm$ 20.4 & 391.7 $\pm$ 9.7 & 304.1 $\pm$ 23.0 & 330.8 $\pm$ 16.0 & 330.9 $\pm$ 8.4 & 540.4 $\pm$ 33.1  \\

    & Random & 240.2 $\pm$ 8.9 & 173.3 $\pm$ 13.8 & 176.2 $\pm$ 13.1 & 151.2 $\pm$ 11.6 & 172.8 $\pm$ 11.2 & 242.7 $\pm$ 20.1  \\

    & Our attack & 221.1 $\pm$ 11.2 & 90.8 $\pm$ 7.8 & 153.7 $\pm$ 14.0 & 88.4 $\pm$ 6.6 & 106.1 $\pm$ 8.2 & 114.1 $\pm$ 16.2  \\

    \hline
    \multirow{3}{*}{BAT+SP} & No attack & 372.8 $\pm$ 9.8 & 459.7 $\pm$ 14.3 & \textbf{373.3 $\pm$ 24.5} & \textbf{352.8 $\pm$ 4.5} & 335.9 $\pm$ 23.4 & 612.2 $\pm$ 35.6  \\

    & Random & 269.6 $\pm$ 15.1 & 279.4 $\pm$ 14.7 & 190.2 $\pm$ 13.0 & 180.6 $\pm$ 13.8 & \textbf{209.3 $\pm$ 14.8} & 364.9 $\pm$ 18.3  \\

    & Our attack & 197.2 $\pm$ 10.9 & 196.3 $\pm$ 16.1 & 138.0 $\pm$ 14.7 & 98.5 $\pm$ 11.8 & \textbf{107.6 $\pm$ 12.3} & 247.8 $\pm$ 15.3  \\
    \hline
    \multirow{3}{*}{BAT+FCP} & No attack & \textbf{456.9 $\pm$ 27.3} & \textbf{460.4 $\pm$ 9.5} & 353.7 $\pm$ 12.8 & 340.9 $\pm$ 19.0 & \textbf{351.8 $\pm$ 15.5} & \textbf{759.0 $\pm$ 36.8}  \\

    & Random &  \textbf{389.8 $\pm$ 9.2} & \textbf{333.8 $\pm$ 11.8} & \textbf{219.2 $\pm$ 12.5} & \textbf{171.3 $\pm$ 11.5} & 178.6 $\pm$ 13.6 & \textbf{467.7 $\pm$ 24.2}  \\

    & Our attack &  \textbf{308.6 $\pm$ 12.3} & \textbf{206.3 $\pm$ 10.1} & \textbf{178.9 $\pm$ 12.3} & \textbf{120.7 $\pm$ 13.0} & 58.3 $\pm$ 9.3 & \textbf{289.4 $\pm$ 24.3}  \\
    \hline
    \end{tabular}
    }
\end{table*}

\subsection{Defense}
\subsubsection{Experimental setup}
We empirically select the top 5 outputs of our attack algorithm and sample 5 random initial states to form the set of perturbed initial states. 
In the supervised kick-start phase, we collect 20 trajectories to construct a dataset for each agent, and each trajectory has 800 environmental steps. The temperature $T$ is set to $1.5$, the hyperparameter $\alpha$ is set to $0.05$, $\beta$ is set to $1$, and the learning rate is set to $0.001$. The data are split into a training set and a validation set with a proportion of $7: 3$. We train the model for $100$ epochs and save the one with the best performance on the validation set. 
The fine-tuning stage uses similar hyperparameters while only training the agent for $8e6$ time steps, keeping the total computational cost of defense close to the original training process.. 
To comprehensively study the effectiveness of our proposed defense method, we employ several baselines for comparison, as listed in the following:
\begin{itemize}
    \item RADIAL\cite{radial}, a robust DRL framework that introduces an adversarial loss in optimization, which aims to resist adversarial perturbation restricted by the $L_p$-norm. Since no previous work targets environmental perturbations, we regard RADIAL as a representation of the existing robust DRL algorithm. We use RADIAL-PPO to train the self-play agents as a baseline. The perturbation bound $\epsilon$ is set to smoothly increase from 0 to $1/255$.
    \item Extra Training. To eliminate the influence of the difference in experimental settings between the original training and the fine-tuning, we additionally train the DRL agents for more time steps with the same setting as BAT without any defense method.
    \item Diversified Start. A natural and widely adopted way to improve the robustness of agents is to train from a distribution of initial states \cite{yang2022optimal, zhang2018dissection}, which is also applicable to defend against environmental perturbations. For this Baseline, the distribution of initial states is the same as the one used in BAT.
\end{itemize}

\subsubsection{Defense results}
We present the mean scores of models under various attacks and corresponding standard errors in \Cref{tab: dfd_result}, where the best scores are in bold. Specifically, the scores achieved in the standard environments measure the capability of the agents, the random perturbations that have no pertinence to the agents measure their anti-interference performance, and the performance under attacks evaluates the resistance of agents. Generally, for both SP and FCP, our BAT can significantly improve their performance across all situations and preserve considerable capabilities under attack, which strongly demonstrates the effectiveness of our framework. It is surprising that BAT also significantly improves the performance even in the standard environment. Such a result indicates that the improvement of our framework comes from the improvement in the capabilities of agents, instead of increasing or decreasing the confidence of choices. We also notice that BAT + SP performs comparably to BAT + FCP, suggesting that our defense framework can produce diversified enough data for defending against environmental attacks while substantiating that our BAT remains effective in general tasks where no MARL algorithm is applicable. 

Since the extra training is in self-play, it seems to bring the SP and FCP agents into convergence. The performance in unperturbed environments of FCP agents is significantly improved and is comparable to that of SP agents. Nevertheless, they are still vulnerable to environmental perturbations.
In contrast, RADIAL shows resistance to environmental perturbations. However, it has significantly lower performance in standard environments, while the defense performance is still much weaker than our BAT. We deem that existing robustness algorithms designed to resist perturbations restricted by the $L_p$-norm are not suitable for defending against environmental perturbations. 
As the strongest baseline, training with diversified initial states also shows considerable defense performance, which is in line with our analysis that diversified training data can improve the robustness of agents. Nevertheless, our BAT remains outperforming it, especially doing much better in the standard environment. Besides, BAT is an entire post-processing that has no limitation on the primary training, while training with diversified initial states may not be applicable for some algorithms, \eg, algorithms including imitation learning \cite{bcp}. 

Another noteworthy point is that our attack is still significantly more effective than random perturbations. It confirms the ubiquitous power of our attack while implying that there is still room for improving the defense algorithms. 

\section{Conclusion and future work} \label{sec: conclusion}
In this paper, we propose a novel threat model in which the saboteur can perturb the initial environmental state within a semantic budget. We further design an algorithm to perform a non-targeted attack by generating adversarial initial states and propose the BAT framework to enhance the capabilities and robustness of agents with the generated adversarial examples. Extensive experimental results demonstrate that the mainstream DRL methods are vulnerable under our threat model and can be attacked by our algorithm. In contrast, our defense method shows significant effectiveness in resisting environmental perturbations and even improves the capability of agents in the pristine environment and outperforms the baselines, including a representative existing robust reinforcement learning algorithm. 

We preliminarily validate the existence of environmental vulnerability of DRL agents, demonstrate the problems of attack and defense, and verify the effectiveness of our methods. Although our problem definition and methods are general and have the potential for further extension, the validation is currently conducted on the 2D game domain, leaving the generalization to the 3D domain and even the real world as future work.

\bibliographystyle{IEEEtran}
\bibliography{references}

\begin{thebibliography}{10}
\providecommand{\url}[1]{#1}
\csname url@samestyle\endcsname
\providecommand{\newblock}{\relax}
\providecommand{\bibinfo}[2]{#2}
\providecommand{\BIBentrySTDinterwordspacing}{\spaceskip=0pt\relax}
\providecommand{\BIBentryALTinterwordstretchfactor}{4}
\providecommand{\BIBentryALTinterwordspacing}{\spaceskip=\fontdimen2\font plus
\BIBentryALTinterwordstretchfactor\fontdimen3\font minus \fontdimen4\font\relax}
\providecommand{\BIBforeignlanguage}[2]{{%
\expandafter\ifx\csname l@#1\endcsname\relax
\typeout{** WARNING: IEEEtran.bst: No hyphenation pattern has been}%
\typeout{** loaded for the language `#1'. Using the pattern for}%
\typeout{** the default language instead.}%
\else
\language=\csname l@#1\endcsname
\fi
#2}}
\providecommand{\BIBdecl}{\relax}
\BIBdecl

\bibitem{sa-mdp}
H.~Zhang, H.~Chen, C.~Xiao, B.~Li, M.~Liu, D.~Boning, and C.-J. Hsieh, ``Robust deep reinforcement learning against adversarial perturbations on state observations,'' \emph{Advances in Neural Information Processing Systems}, vol.~33, pp. 21\,024--21\,037, 2020.

\bibitem{radial}
T.~Oikarinen, W.~Zhang, A.~Megretski, L.~Daniel, and T.-W. Weng, ``Robust deep reinforcement learning through adversarial loss,'' in \emph{Advances in Neural Information Processing Systems}, 2021.

\bibitem{zhang2021robust}
H.~Zhang, H.~Chen, D.~Boning, and C.-J. Hsieh, ``Robust reinforcement learning on state observations with learned optimal adversary,'' in \emph{International Conference on Learning Representation (ICLR)}, 2021.

\bibitem{liang2022efficient}
Y.~Liang, Y.~Sun, R.~Zheng, and F.~Huang, ``Efficient adversarial training without attacking: Worst-case-aware robust reinforcement learning,'' in \emph{Advances in Neural Information Processing Systems}, 2022.

\bibitem{lee2020spatiotemporally}
X.~Y. Lee, S.~Ghadai, K.~L. Tan, C.~Hegde, and S.~Sarkar, ``Spatiotemporally constrained action space attacks on deep reinforcement learning agents,'' in \emph{Proceedings of the AAAI conference on artificial intelligence}, vol.~34, no.~04, 2020, pp. 4577--4584.

\bibitem{gunn2022adversarial}
S.~Gunn, D.~Jang, O.~Paradise, L.~Spangher, and C.~J. Spanos, ``Adversarial poisoning attacks on reinforcement learning-driven energy pricing,'' in \emph{Proceedings of the 9th ACM International Conference on Systems for Energy-Efficient Buildings, Cities, and Transportation}, 2022, pp. 262--265.

\bibitem{panagiota2020trojdrl}
K.~Panagiota, W.~Kacper, S.~Jha, and L.~Wenchao, ``Trojdrl: Trojan attacks on deep reinforcement learning agents. in proc. 57th acm/ieee design automation conference (dac), 2020, march 2020,'' in \emph{Proc. 57th ACM/IEEE Design Automation Conference (DAC), 2020}, 2020.

\bibitem{gleave2019adversarial}
A.~Gleave, M.~Dennis, C.~Wild, N.~Kant, S.~Levine, and S.~Russell, ``Adversarial policies: Attacking deep reinforcement learning,'' in \emph{International Conference on Learning Representations}, 2020.

\bibitem{lin2020robustness}
J.~Lin, K.~Dzeparoska, S.~Q. Zhang, A.~Leon-Garcia, and N.~Papernot, ``On the robustness of cooperative multi-agent reinforcement learning,'' in \emph{2020 IEEE Security and Privacy Workshops (SPW)}.\hskip 1em plus 0.5em minus 0.4em\relax IEEE, 2020, pp. 62--68.

\bibitem{guo2022towards}
J.~Guo, Y.~Chen, Y.~Hao, Z.~Yin, Y.~Yu, and S.~Li, ``Towards comprehensive testing on the robustness of cooperative multi-agent reinforcement learning,'' in \emph{Proceedings of the IEEE/CVF Conference on Computer Vision and Pattern Recognition}, 2022, pp. 115--122.

\bibitem{bcp}
M.~Carroll, R.~Shah, M.~K. Ho, T.~Griffiths, S.~Seshia, P.~Abbeel, and A.~Dragan, ``On the utility of learning about humans for human-ai coordination,'' in \emph{Advances in neural information processing systems}, vol.~32, 2019.

\bibitem{fcp}
D.~Strouse, K.~R. McKee, M.~Botvinick, E.~Hughes, and R.~Everett, ``Collaborating with humans without human data,'' in \emph{Advances in Neural Information Processing Systems}, vol.~34, 2021, pp. 14\,502--14\,515.

\bibitem{zhao2022coordination}
M.~Zhao, R.~Simmons, and H.~Admoni, ``Coordination with humans via strategy matching,'' in \emph{2022 IEEE/RSJ International Conference on Intelligent Robots and Systems (IROS)}.\hskip 1em plus 0.5em minus 0.4em\relax IEEE, 2022, pp. 9116--9123.

\bibitem{yu2023learning}
C.~Yu, J.~Gao, W.~Liu, B.~Xu, H.~Tang, J.~Yang, Y.~Wang, and Y.~Wu, ``Learning zero-shot cooperation with humans, assuming humans are biased,'' in \emph{International Conference on Learning Representations}, 2023.

\bibitem{yan2023efficient}
X.~Yan, J.~Guo, X.~Lou, J.~Wang, H.~Zhang, and Y.~Du, ``An efficient end-to-end training approach for zero-shot human-ai coordination,'' in \emph{NeurIPS 2023}, 2023.

\bibitem{szegedy2013intriguing}
C.~Szegedy, W.~Zaremba, I.~Sutskever, J.~Bruna, D.~Erhan, I.~Goodfellow, and R.~Fergus, ``Intriguing properties of neural networks,'' in \emph{International Conference on Learning Representations, {ICLR}}, 2014.

\bibitem{su2019one}
J.~Su, D.~V. Vargas, and K.~Sakurai, ``One pixel attack for fooling deep neural networks,'' \emph{IEEE Transactions on Evolutionary Computation}, vol.~23, no.~5, pp. 828--841, 2019.

\bibitem{huang2017adversarial}
S.~Huang, N.~Papernot, I.~Goodfellow, Y.~Duan, and P.~Abbeel, ``Adversarial attacks on neural network policies,'' \emph{arXiv preprint arXiv:1702.02284}, 2017.

\bibitem{behzadan2017vulnerability}
V.~Behzadan and A.~Munir, ``Vulnerability of deep reinforcement learning to policy induction attacks,'' in \emph{International Conference on Machine Learning and Data Mining in Pattern Recognition}.\hskip 1em plus 0.5em minus 0.4em\relax Springer, 2017, pp. 262--275.

\bibitem{lin2017tactics}
Y.-C. Lin, Z.-W. Hong, Y.-H. Liao, M.-L. Shih, M.-Y. Liu, and M.~Sun, ``Tactics of adversarial attack on deep reinforcement learning agents,'' in \emph{Proceedings of the 26th International Joint Conference on Artificial Intelligence}, 2017, pp. 3756--3762.

\bibitem{weng2020toward}
T.-W. Weng, K.~D. Dvijotham, J.~Uesato, K.~Xiao, S.~Gowal, R.~Stanforth, and P.~Kohli, ``Toward evaluating robustness of deep reinforcement learning with continuous control,'' in \emph{International Conference on Learning Representations}, 2020.

\bibitem{tekgul2022real}
B.~G. Tekgul, S.~Wang, S.~Marchal, and N.~Asokan, ``Real-time adversarial perturbations against deep reinforcement learning policies: attacks and defenses,'' in \emph{European Symposium on Research in Computer Security}.\hskip 1em plus 0.5em minus 0.4em\relax Springer, 2022, pp. 384--404.

\bibitem{buddareddygari2022targeted}
P.~Buddareddygari, T.~Zhang, Y.~Yang, and Y.~Ren, ``Targeted attack on deep rl-based autonomous driving with learned visual patterns,'' in \emph{2022 International Conference on Robotics and Automation (ICRA)}.\hskip 1em plus 0.5em minus 0.4em\relax IEEE, 2022, pp. 10\,571--10\,577.

\bibitem{schott2022improving}
L.~Schott, H.~Hajri, and S.~Lamprier, ``Improving robustness of deep reinforcement learning agents: Environment attack based on the critic network,'' in \emph{2022 International Joint Conference on Neural Networks (IJCNN)}.\hskip 1em plus 0.5em minus 0.4em\relax IEEE, 2022, pp. 1--8.

\bibitem{pan2022characterizing}
X.~Pan, C.~Xiao, W.~He, S.~Yang, J.~Peng, M.~Sun, M.~Liu, B.~Li, and D.~Song, ``Characterizing attacks on deep reinforcement learning,'' in \emph{Proceedings of the 21st International Conference on Autonomous Agents and Multiagent Systems}, 2022, pp. 1010--1018.

\bibitem{uesato2019rigorous}
J.~Uesato, A.~Kumar, C.~Szepesvari, T.~Erez, A.~Ruderman, K.~Anderson, N.~Heess, P.~Kohli \emph{et~al.}, ``Rigorous agent evaluation: An adversarial approach to uncover catastrophic failures,'' in \emph{International Conference on Learning Representations}, 2019.

\bibitem{panda2018scalable}
S.~Panda and Y.~Vorobeychik, ``Scalable initial state interdiction for factored mdps,'' in \emph{International Joint Conference on Artificial Intelligence}, 2018.

\bibitem{pinto2017robust}
L.~Pinto, J.~Davidson, R.~Sukthankar, and A.~Gupta, ``Robust adversarial reinforcement learning,'' in \emph{International Conference on Machine Learning}.\hskip 1em plus 0.5em minus 0.4em\relax PMLR, 2017, pp. 2817--2826.

\bibitem{pattanaik2018robust}
A.~Pattanaik, Z.~Tang, S.~Liu, G.~Bommannan, and G.~Chowdhary, ``Robust deep reinforcement learning with adversarial attacks,'' in \emph{Proceedings of the 17th International Conference on Autonomous Agents and MultiAgent Systems}, 2018, pp. 2040--2042.

\bibitem{wu2022robust}
J.~Wu and Y.~Vorobeychik, ``Robust deep reinforcement learning through bootstrapped opportunistic curriculum,'' in \emph{International Conference on Machine Learning}.\hskip 1em plus 0.5em minus 0.4em\relax PMLR, 2022, pp. 24\,177--24\,211.

\bibitem{fischer2019online}
M.~Fischer, M.~Mirman, S.~Stalder, and M.~Vechev, ``Online robustness training for deep reinforcement learning,'' \emph{arXiv preprint arXiv:1911.00887}, 2019.

\bibitem{lutjens2020certified}
B.~L{\"u}tjens, M.~Everett, and J.~P. How, ``Certified adversarial robustness for deep reinforcement learning,'' in \emph{Conference on Robot Learning}.\hskip 1em plus 0.5em minus 0.4em\relax PMLR, 2020, pp. 1328--1337.

\bibitem{everett2021certifiable}
M.~Everett, B.~L{\"u}tjens, and J.~P. How, ``Certifiable robustness to adversarial state uncertainty in deep reinforcement learning,'' \emph{IEEE Transactions on Neural Networks and Learning Systems}, 2021.

\bibitem{di2022goal}
L.~L. Di~Langosco, J.~Koch, L.~D. Sharkey, J.~Pfau, and D.~Krueger, ``Goal misgeneralization in deep reinforcement learning,'' in \emph{International Conference on Machine Learning}.\hskip 1em plus 0.5em minus 0.4em\relax PMLR, 2022, pp. 12\,004--12\,019.

\bibitem{hinton2015distilling}
G.~Hinton, O.~Vinyals, and J.~Dean, ``Distilling the knowledge in a neural network,'' \emph{arXiv preprint arXiv:1503.02531}, 2015.

\bibitem{papernot2016distillation}
N.~Papernot, P.~McDaniel, X.~Wu, S.~Jha, and A.~Swami, ``Distillation as a defense to adversarial perturbations against deep neural networks,'' in \emph{2016 IEEE symposium on security and privacy (SP)}.\hskip 1em plus 0.5em minus 0.4em\relax IEEE, 2016, pp. 582--597.

\bibitem{todorov2012mujoco}
E.~Todorov, T.~Erez, and Y.~Tassa, ``Mujoco: A physics engine for model-based control,'' in \emph{2012 IEEE/RSJ international conference on intelligent robots and systems}.\hskip 1em plus 0.5em minus 0.4em\relax IEEE, 2012, pp. 5026--5033.

\bibitem{cobbe2020leveraging}
K.~Cobbe, C.~Hesse, J.~Hilton, and J.~Schulman, ``Leveraging procedural generation to benchmark reinforcement learning,'' in \emph{International conference on machine learning}.\hskip 1em plus 0.5em minus 0.4em\relax PMLR, 2020, pp. 2048--2056.

\bibitem{schulman2017proximal}
J.~Schulman, F.~Wolski, P.~Dhariwal, A.~Radford, and O.~Klimov, ``Proximal policy optimization algorithms,'' \emph{arXiv preprint arXiv:1707.06347}, 2017.

\bibitem{yang2022optimal}
M.~Yang, M.~Carroll, and A.~Dragan, ``Optimal behavior prior: Data-efficient human models for improved human-ai collaboration,'' \emph{arXiv preprint arXiv:2211.01602}, 2022.

\bibitem{zhang2018dissection}
A.~Zhang, N.~Ballas, and J.~Pineau, ``A dissection of overfitting and generalization in continuous reinforcement learning,'' \emph{arXiv preprint arXiv:1806.07937}, 2018.

\end{thebibliography}

\end{document}